\documentclass[journal ]{new-aiaa}
\usepackage[utf8]{inputenc}
\usepackage{textcomp}

\usepackage{graphicx}
\usepackage{graphicx} 
\usepackage{textcomp}
\usepackage{caption,subcaption}
\usepackage{geometry}
\usepackage{placeins}
\usepackage{amsmath}
\usepackage[version=4]{mhchem}
\usepackage{siunitx}
\usepackage{longtable,tabularx}
\usepackage[ruled,vlined]{algorithm2e}
\usepackage{makecell}
\newcommand*{\QEDA}{\null\nobreak\hfill\ensuremath{\blacksquare}}%

\title{Safe and Scalable Real-Time Trajectory Planning Framework for Urban Air Mobility}

\author{Abenezer G. Taye\footnote{Graduate Student, Department of Mechanical \& Aerospace Engineering, abenezertaye@gwu.edu,  AIAA Student Member.}} 
\affil{George Washington University, Washington, DC, 20052, USA}

\author{Roberto Valenti\footnote{Research Scientist, MathWorks Advanced Research and Technology Office, rvalenti@mathworks.com}} 
\affil{MathWorks, Natick, MA 01760, USA}

\author{Akshay Rajhans\footnote{Research Scientist, MathWorks Advanced Research and Technology Office, arajhans@mathworks.com }} 
\affil{MathWorks, Natick, MA 01760, USA}

\author{Anastasia Mavrommati\footnote{Research Scientist, MathWorks Advanced Research and Technology Office, amavromm@mathworks.com }} 
\affil{MathWorks, Natick, MA 01760, USA}

\author{Pieter J. Mosterman\footnote{Research Scientist, MathWorks Advanced Research and Technology Office }} 
\affil{MathWorks, Natick, MA 01760, USA}

\author{Peng Wei\footnote{Associate Professor, Department of Mechanical \& Aerospace Engineering, pwei@gwu.edu,  AIAA Senior Member.}} 
\affil{George Washington University, Washington, DC, 20052, USA}
\begin{document}

\maketitle

\begin{abstract}
This paper presents a real-time trajectory planning framework for Urban Air Mobility (UAM) that is both safe and scalable. The proposed framework employs a decentralized, free-flight concept of operation in which each aircraft independently performs separation assurance and conflict resolution, generating safe trajectories by accounting for the future states of nearby aircraft. The framework consists of two main components: a data-driven reachability analysis tool and an efficient Markov Decision Process (MDP) based decision maker. The reachability analysis over-approximates the reachable set of each aircraft through a discrepancy function learned online from simulated trajectories. The decision maker, on the other hand, uses a 6-degrees-of-freedom guidance model of fixed-wing aircraft to ensure collision-free trajectory planning. Additionally, the proposed framework incorporates reward shaping and action shielding techniques to enhance safety performance. The proposed framework is evaluated through simulation experiments involving up to 32 aircraft in a UAM setting, with performance measured by the number of Near Mid Air Collisions (NMAC) and computational time. The results demonstrate the safety and scalability of the proposed framework.
\end{abstract}

\section{Introduction}
\label{sec:introduction}

\subsection{Motivation}
Urban Air Mobility (UAM) is a novel concept in which partially or fully autonomous air vehicles transport passengers and cargo in dense urban environments. This technology aims to provide a safe, efficient, and accessible on-demand air transportation system \cite{patterson_2018}, offering an alternative to traditional ground-based transportation methods. Furthermore, as the technology advances, it will connect urban centers to outlying areas, expanding the reach of metropolitan regions. 

UAM operation is a multi-agent safety-critical application that requires the simultaneous consideration of safety and scalability as primary design considerations. Thus, a UAM trajectory planning framework needs to generate trajectories efficiently while ensuring compliance with system safety requirements. These two problems --- developing a scalable trajectory planner and safety verification of autonomous systems --- are fundamentally challenging in and of themselves and are often addressed independently in the literature. However, in the context of UAM, both must be considered simultaneously. 

The task of guaranteeing the safe operation of autonomous systems is often called verification and validation. Several approaches to verification and validation have been proposed in the literature. These approaches can be broadly classified as formal methods and sampling-based approaches. Sampling-based approaches involve generating a finite number of scenarios to assess the performance of a system. Hence, they have the advantage of being easier to implement and evaluate the performance of an autonomous system. However, they can not account for all possible behaviors of the system, which is an essential element in verification and validation. As a result, formal methods, which can capture all possible behaviors of the system, have gained significant research attention in recent years.

\subsection{Related Work}
\label{sec:related_work}

Organizations such as NASA\texttrademark, Uber\texttrademark, and {Airbus\texttrademark} have been exploring the use of vertical takeoff and landing (VTOL) aircraft for UAM \cite{gipson_2017,uber_elevate, holden2016fast, airbus_2018,airbus_2019}. The UAM concept envisions the use of VTOL aircraft departing and arriving at small-scale airports known as vertiports.

An unstructured airspace approach known as \textit{“free flight”} has been proposed as a solution to the ongoing congestion of the current Air Traffic Control (ATC) system. Studies have demonstrated that free flight with airborne separation can handle a higher traffic density \cite{hoekstra2002designing, bilimoria2000performance}, and bring fuel and time efficiency \cite{valenti2001cost}. Under this approach, each aircraft performs separation assurance and conflict resolution. Tomlin \textit{et al.} \cite{tomlin1998conflict} stated that free flight is potentially feasible due to enabling technologies such as Global Positioning Systems (GPS), data link communications like Automatic Dependence Surveillance-Broadcast (ADS-B) \cite{kahne1996air}, Traffic Alert and Collision Avoidance Systems (TCAS) \cite{harman1989tcas}, but would require robust onboard computation.

The literature on multi-agent trajectory planning algorithms is extensive and can broadly be classified as centralized and decentralized methods. In centralized methods, the state of each aircraft, obstacles, trajectory constraints, and the terminal area’s state are observable to the controller via sensors, radar, etc., and a central supervising controller resolves conflicts between aircraft. The central controller precomputes trajectories for all aircraft before flight, typically by formulating the problem in an optimal control framework and solving the problem with various methods; examples are: semidefinite programming \cite{frazzoli2001resolution}, nonlinear programming \cite{raghunathan2004dynamic, enright1992discrete}, mixed-integer linear programming \cite{schouwenaars2001mixed, richards2002aircraft, pallottino2002conflict, vela2009mixed}, mixed-integer quadratic programming \cite{mellinger2012mixed}, sequential convex programming \cite{augugliaro2012generation, morgan2014model}, second-order cone programming \cite{acikmese2007convex}, evolutionary techniques \cite{delahaye2010aircraft, cobano2011path}, reinforcement learning \cite{razzaghi2211survey}, and particle swarm optimization \cite{pontani2010particle}. One common thread among centralized approaches is that to pursue a global optimum, they must consider each aircraft and obstacle in space, leading to scalability issues with a large number of aircraft and obstacles. In addition, as new aircraft enter the scene, centralized algorithms typically need to recompute part or all of the problem to arrive at a new global optimum.

On the other hand, decentralized methods scale better with the number of aircraft and objects in the system but typically cannot obtain globally optimal solutions. Furthermore, decentralized methods may be more robust than centralized approaches \cite{pallottino2006probabilistic} because they are not generally prone to a single point of failure. In decentralized systems, each aircraft resolves conflicts locally, and the underlying method can be considered either cooperative or non-cooperative. Computational scalability and solution quality or optimality are significant design trade-offs between centralized and decentralized trajectory planning strategies. In \cite{bertram2020distributed}, we proposed a Markov Decision Process (MDP) based decentralized UAM trajectory planning algorithm that is highly scalable. The algorithm operates in a free-flight manner. This study is extended by incorporating an online safety verification module that enables the trajectory planner to generate safe trajectories.

From a safety verification standpoint, trajectory planning of autonomous systems has recently been studied in two main directions: \textit{design-then-verify} and \textit{verify-while-design}. \textit{Design-then-verify} is a commonly used approach where the task of trajectory planning is performed first; then, the system is evaluated using different verification tools to determine whether it satisfies the safety requirements \cite{duggirala2015c2e2,larsen1997uppaal}. However, this approach is computationally inefficient and often fails to give the necessary guarantees \cite{wang2022design}. On the other hand, the \textit{verify-while-design} approach, also known as correct-by-construction, integrates the verification process into the control design in a closed-loop manner \cite{fan2021controller,fisac2015reach}. Thus the approach becomes computationally efficient and enables the system to satisfy the safety requirements by its very nature.

In this study, we adopted the \textit{verify-while-design} approach to synthesize each aircraft's trajectory online formally. An efficient reachability analysis module that explores all possible behaviors of an aircraft has been used to satisfy the reach-avoid property of the system. Several reachability analysis formulations of a dynamical system have been proposed in the literature. These methods include Hamilton-Jacobi-based reachability analysis formulations \cite{bansal2017hamilton}, CORA \cite{althoff2015introduction}, SpaceEx \cite{frehse2011spaceex}, and Flow$^{*}$ \cite{chen2013flow}. Although these approaches provide formal soundness guarantees, they are computationally expensive. Hence, they can not be used online in the presence of many aircraft. In this study, to over-approximate the reachable set of an aircraft, we implemented a sensitivity analysis-based approach from DryVR \cite{10.1007/978-3-319-63387-9_22}. DryVR has been demonstrated to be highly scalable and recently implemented in \cite{hsieh2021skytrakx} to generate a safe operation volume for unmanned aircraft systems (UAS) traffic management. The reachability analysis module, then, is integrated with our previously developed MDP-based trajectory planner \cite{bertram2020distributed} to guide the motion of multiple UAM vehicles between vertiports.

\subsection{Overview of the Paper}
We presented a preliminary version of this paper at the AIAA Aviation 2022 conference \cite{taye2022reachability}. The contribution of this paper is threefold. First, we formulate the safe multi-agent trajectory planning problem using a reachability analysis module and an MDP-based decision-maker. To achieve overall system scalability, highly scalable approaches have been employed for both components. Second, we propose a reward-shaping mechanism that enhances the safety properties of the trajectory planner by modifying the properties of its reward function. Third, we propose an action shielding strategy that further enhances the safety properties of the system by filtering out actions that lead to unsafe states.

This paper is organized as follows: In Section \ref{sec:related_work}, we reviewed previous works related to the problem at hand. Section \ref{sec:problem_formulation} outlines the problem, and section \ref{sec:modeling} presents the mathematical formulation of the two main components, the MDP and reachability analysis. We also provide an overview of the proposed trajectory planning framework, including the role of each component in the trajectory planning procedure. In Section \ref{sec:results}, we discuss the implemented UAM scenario and present the results for the nominal trajectory planner (without any safety reinforcement) and the two other approaches proposed to improve the safety of the trajectory planner, namely, action shielding and reward shaping. Finally, in Section \ref{sec:conclusion}, we provide the conclusion of this work.

\section{Problem Formulation}
\label{sec:problem_formulation}

\subsection{Problem Description}
\begin{figure}[h]
\begin{center}
\includegraphics[width=0.7\textwidth]{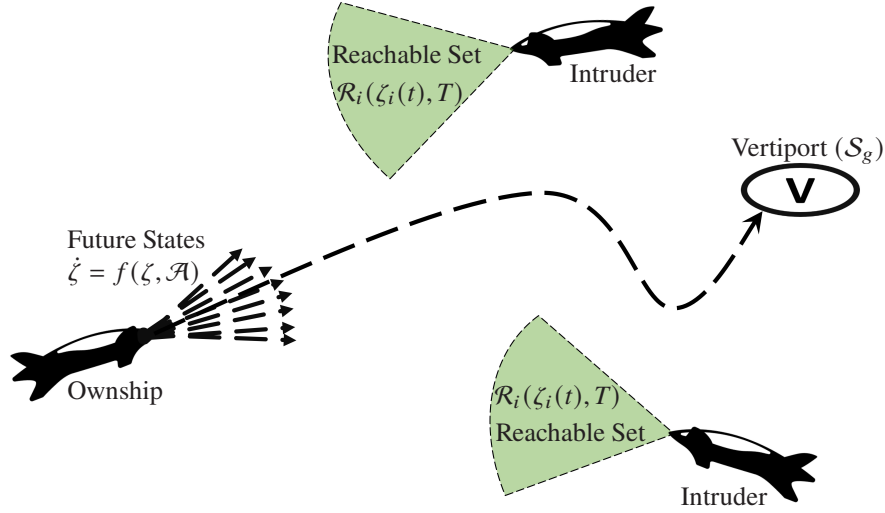}
\put(-68,5){Intruder}
\put(-110,165){Intruder}
\put(-49,137){Vertiport $(\mathcal{S}_g)$}
\put(-300,102){Future States}
\put(-300,90){$\dot{\zeta} = f(\zeta,\mathcal{A})$}
\put(-300,45){Ownship}
\put(-200,172){Reachable Set}
\put(-198,158){$\mathcal{R}_i(\zeta_i(t),T)$}
\put(-141,43){ $\mathcal{R}_i(\zeta_i(t),T)$}
\put(-138,29){Reachable Set}

\caption{\textbf{Working principle of the proposed trajectory planner. The proposed trajectory planner operates based on a combination of a reachability analysis module and a Markov Decision Process (MDP) based decision-making scheme. The reachability analysis module is used to determine the possible future states of the intruder aircraft. The MDP-based decision-making scheme, in turn, is used to guide the aircraft towards its assigned destination vertiport while avoiding entering the reachable sets of nearby intruders.}}

\label{fig:MDP_figure}
\end{center}
\end{figure}

This study aims to address the problem of developing a UAM trajectory planning framework that is computationally efficient and guarantees the safe navigation of UAM aircraft. As shown in Figure \ref{fig:MDP_figure}, the two main components of the proposed framework are the MDP-based trajectory planner and a reachability analysis module, which the trajectory planner utilizes to gather information about the future states of the aircraft. The approaches we used to formulate the trajectory planning problem and compute the reachable sets of the aircraft are proven to be highly scalable \cite{bertram2020distributed}\cite{10.1007/978-3-319-63387-9_22}. Adopting such formulations makes the developed UAM trajectory planning framework computationally efficient. Furthermore, the algorithm allows each aircraft to make its own decisions in a distributed manner using inputs from sensors such as radar, LIDAR, or systems such as ADS-B.

\subsection{Aircraft Dynamics}\label{subsection:vehicle}
\vspace{0.1cm}
The aircraft model used in this paper is based on a 6-DOF kinematic guidance model formulation proposed in \cite{beard2012small}. The original guidance model contains certain wind-related parameters. However, since we are not considering the presence of wind in this study, we used a simplified model given in Equation \ref{guidance}, where $\Dot{x}, \Dot{y}, \Dot{z}$ are north, east, and down velocities of the aircraft with respect to the inertial reference frame. $\gamma$ is the flight-path angle, and $V$ is the speed of the aircraft. $\phi$, $\chi$, and $\psi$ represent the roll, course, and heading angles, respectively. $b_\gamma$, $b_{V}$, and $b_\phi$ are positive constants that depend on the implementation of the autopilot and the state estimation schemes. The superscript $*^c$ as in $\gamma^c, V^c$, and $\phi^c$ denotes the commanded values given to the autopilot. 

\begin{equation}\label{guidance}
    \begin{cases}
      \Dot{x} =& V \cos \psi \cos \gamma \\
      \Dot{y} =& V \sin \psi \cos \gamma \\
      \Dot{z} =& V \sin \gamma \\
      \Dot{\chi} =& \frac{g}{V} \tan \phi \cos(\chi - \psi)\\
      \Dot{\gamma} =& b_\gamma (\gamma^c - \gamma)\\
      \Dot{V} =& b_{V} (V^c - V)\\
      \Dot{\phi} =& b_{\phi} (\phi^c - \phi)\\
    \end{cases} 
\end{equation}

\section{Methodology}
\label{sec:modeling}

\subsection{Markov Decision Process Formulation} 
In this paper, we formulate the aircraft trajectory planning problem as a Markov decision process (MDP), where the state transitions will be governed by the vehicle dynamics described in Section \ref{subsection:vehicle}. MDPs are formulated as the tuple $(s_t,a_t,r_t,t)$ where $s_t \in S$ is the state at a given time $t$ within the state space $S$. $a_t \in \mathcal{A}$ denotes the action taken by the agent at time $t$ from the action set $\mathcal{A}$. $r_t$ is the reward received by the agent as a result of taking action $a_t$ from $s_t$ and arriving at $s_{t+1}$, and $T(s_t, a, s_{t+1})$ is a transition function that describes the dynamics of the environment and capture the probability $p(s_{t+1}|s_t, a_t)$ of transitioning to a state $s_{t+1}$ given the action $a_t$ taken from state $s_t$.

A policy $\pi$ can map each state $ s \in S$ to action $a \in \mathcal{A}$. From a given policy $\pi \in \Pi$, a value function $V^{\pi}(S)$ can be computed that represents the expected return that will be obtained within the environment by following the policy $\pi$. The solution of an MDP is the optimal policy $\pi^*$, which defines the optimal action $a^* \in \mathcal{A}$ that can be taken from each state $s \in S$ to maximize the expected return. From this optimal policy $\pi^*$, the optimal value function $V^*(s)$ can be computed, which describes the maximum expected value obtained from each state $s \in S $. Furthermore, from the optimal value function $V^*(s)$, the optimal policy $\pi^*$ can also easily be recovered.
\subsubsection{State Space}
The environment is a continuous state space placed on a spherical volume of $15km$ radius. Given the dynamics of an aircraft: 

\begin{equation}\label{eqn_1}
    \dot{\zeta}(t) = f(\zeta(t),u(t)),
\end{equation}
where, $f: \mathbb{R}^n \times \mathbb{R} \rightarrow \mathbb{R}^n $ is a continuous function. $\zeta$ denotes the aircraft states, which includes the $x,y,z$ positions, heading angle $\psi$, the flight path angle $\gamma$, the course $\chi$, the roll angle $\phi$, and the speed $V$. The trajectory of an aircraft $\xi: \mathbb{R}^n \times \mathbb{R}_{\geq 0} \rightarrow \mathbb{R}^n$ is the solution to the differential equation (\ref{eqn_1}). It represents how the state variables of the aircraft evolve through time. For a given initial set $x_0 \in \mathbb{R}^n$, the state of the system at time $t$ is $\xi(\zeta_0,t) = \zeta(t)$. The control input $u(t)$ is comprised of the thrust $n_x$, the rate of change of angle of attack $\Dot{\alpha}$, and the rate of change of the roll angle $\Dot{\phi} $.  In addition, a single state in the state space ($s_{o}$) contains all the states of an aircraft ($\zeta$) and the states of every other aircraft denoted as $f_j$, $\forall j \in J$, where $J$ represents a set containing all aircraft in the system except the ownship. Thus, we can define $s_{o}$ as $s_{o} = [\zeta, f_{1}, ..., f_{j}]$. 

\subsubsection{Action Space}  
The action space of the MDP is composed of the individual action spaces of the three inputs: the commanded flight-path angle ($\gamma^c$), the commanded roll angle $\phi^c$, and the commanded airspeed ($V^c$). The action space of $V^c$ is composed of $10$ linearly spaced discrete values between $25m/s$ and $70m/s$. The minimum speed of $25m/s$ is chosen based on the stall speed performance of the aircraft \cite{jackson2004all}. On the other hand, the action spaces of $\gamma^c$ and $\phi^c$ are discrete sets of actions sampled from a logarithm function through the range of each input. Such an action space enables one to take more control actions when the inputs are near zero, and coarse control actions as the aircraft gets further away from its trajectory. As a result, fine control actions can be taken when a small correcting action to adjust small deviations from the trajectory is desired, and large control actions can be taken when a significant change in the course of the aircraft trajectory is desired. Consequently, the inputs of $\gamma^c$ and $\phi^c$ are logarithmically spaced within a range of $15$ input values.

The logarithmically spaced input set in degree is computed as follows:
\begin{equation}
\gamma^c = [-19.99, -16.24, -12.66, -9.26, -6.02, -2.94, -0.01, 0, 0.01, 2.94, 6.02, 9.26, 12.66, 16.24, 19.99]
\end{equation}
\begin{equation}
\indent \phi^c = [-19.99, -16.24, -12.66, -9.26, -6.02, -2.94, -0.01, 0, 0.01, 2.94, 6.02, 9.26, 12.66, 16.24, 19.99]
\end{equation}

Finally, the joint action space becomes:

\begin{equation}
 \mathcal{A} = \{  \gamma^c, \phi^c, V_a^c \}. 
\end{equation}

\subsubsection{Reward Function} \label{ssec:reward}
The reward function is the primary mechanism we use to control the behavior of an MDP agent's behavior. A reward function $R(s_t,a_t,s_{t+1})$ represents the reward that an agent, currently at $s_t$, collects after taking a control action $a_t$ and arriving at $s_{t+1}$. In this work, we have utilized both positive and negative rewards, as depicted in Table \ref{table_2}, to guide the aircraft to their destination while avoiding possible collision with other nearby aircraft. A negative reward function that scales linearly inside the reach set of the intruder aircraft is employed instead of a constant negative reward value to prevent closer proximity between aircraft. 
\vspace{0.5cm}
\begin{table}[!h]
\begin{center}
\begin{tabular}{|| c c c c c ||}
\hline 
 Reward source & Reward magnitude & Location & Decay factor & Description \\ [0.5ex]
 \hline \hline
 Intruder aircraft & $-(100t + 500)$ & Inside reachable-set of intruder   & $0.97$ & Collision avoidance \\
\hline
 Destination & $200$ & Manually placed & $0.999$ & Vertiport attraction \\
\hline
\end{tabular}
\caption{ \textbf{Reward function for each aircraft}}
\label{table_2}
\end{center}
\end{table}
\vspace{-0.5cm}

\subsubsection{Value Function}
Once the MDP is formulated as a tuple of $(s_t,a_t,r_t)$, we need to solve the formulated MDP to arrive at the optimal solution. The state-value function $(V(s))$ is used to determine the expected reward at each future state, allowing for selecting the optimal state. The specific value function structure adopted from \cite{9681334} is defined for deterministic terminating MDPs. The methods and proofs for computing the state-value function are detailed in the full paper; only a summary of the computation process is presented here.
\begin{equation}\label{value_function}
    V(s) = V^+(s) + V^-(s),
\end{equation}
where $V^+(s)$ and $V^-(s)$ are the state-wise positive and negative value functions, respectively.  $V^+(s)$ and $V^-(s)$ are defined as follows:
\begin{equation}
     V^+(s) = \underset{i}{\mathrm{max}}P_i^{+}(s), \; \forall i = \{1, \dots , N^+ \},
\end{equation}
\begin{equation}
     V^-(s) = \underset{j}{\mathrm{min}}P_j^{-}(s), \; \forall j = \{1, \dots , N^- \},
\end{equation}
where $P_i^+$ and $P_j^-$ are the positive and negative peaks created by a reward source $R_i$ and are computed as:
\begin{equation}
    P^+_i(s) = \kappa^{\delta(s,s_i)} \cdot r_i \text{ and } P^-_j(s) = \kappa^{\delta(s,s_i)} \cdot r_i,
\end{equation}
where, $\kappa$ is the discounting factor and $\delta(s,s_i)$ is the distance between the current state $s$ and the reward source $s_i$. 

\subsection{Reachability Analysis} \label{Rechability_subsection}
One of the critical components of the present trajectory planning scheme is computing a reachable set for each nearby intruder aircraft. In this study, the concept of discrepancy function is adopted from \cite{10.1007/978-3-319-63387-9_22} to formulate the reachability analysis problem. This section summarizes discrepancy functions and how they can be used to compute the reachable set of a dynamical system.

A \textit{discrepancy function} is a continuous function primarily used to measure the convergence or divergence nature of trajectories formally \cite{duggirala2013verification}. Hence, it generates the over-approximation of the reachable set by providing the upper and lower bounds of the trajectories. In \cite{duggirala2013verification}, it has been demonstrated that discrepancy functions are generalizations of other well-known proof certificates, such as Contraction metrics and Incremental Lyapunov functions. A discrepancy function $\beta: \mathbb{R}^n \times \mathbb{R}^n \times \mathbb{R}_{\geq0}  \rightarrow \mathbb{R}_{\geq0}$ has two requirements:

\begin{enumerate}
\item $\beta$ upper bounds the distance between the trajectories, 
\vspace{-0.5cm}
\begin{equation}\label{first_req}
    \Vert \xi(\zeta_0,t) - \xi(\zeta'_0,t) \Vert \leq \beta(\zeta_0,\zeta'_0,t),  
\end{equation}
where, $\xi(\zeta_0,t)$ and $\xi(\zeta'_0,t)$ represent any pair of trajectories with initial conditions $\zeta_0$ and $\zeta'_0$, respectively. 

\item $\beta$ converges to zero as the initial states of the trajectories converge.
\vspace{-0.5cm}
\begin{equation}
    \text{for any $t$, as } \zeta_0 \rightarrow \zeta'_0, \;  \beta(\cdot,\cdot,t) \rightarrow 0.
\end{equation}

\end{enumerate}
The first requirement expresses $\beta$ as a function of the initial conditions of any two trajectories and the elapsed time. It upper bounds the distance between the trajectories at any time so that every possible state of the system is represented in the reachable set. On the other hand, the second requirement is used to keep the over-approximation error low.

 There are methods developed in the literature to compute $\beta$ from differential equations \cite{fan2015bounded}. However, in this study, we use a tool known as DryVR \cite{10.1007/978-3-319-63387-9_22} that formulates the problem of finding the discrepancy function as a problem of learning linear separator to achieve high computational efficiency. The learning linear separator approach does not depend on the system's dynamics and uses a few simulations to arrive at a discrepancy function with probabilistic correctness guarantees.

The discrepancy function adopted in DryVR is an exponential function that grows and shrinks with time and has a general form: 
\begin{equation}\label{disc_general}
    \beta( u,v,t) = \Vert u - v \Vert Ke^{ \hat{\gamma} t},
\end{equation}
where $K$ and $\hat{\gamma}$ (we write $\hat{\gamma}$ to distinguish from $\gamma$, which is the flight path angle) are constants that govern the behavior of the exponential function, and we learn them using the learning linear separator approach.

Considering Equation (\ref{disc_general}) and the first requirement of a discrepancy function in Equation (\ref{first_req}):

\begin{equation}\label{eqn_19}
    \Vert \xi(\zeta_0,t) - \xi(\zeta'_0,t)\Vert\leq \Vert \zeta_0 - \zeta'_0 \Vert Ke^{\hat{\gamma} t},\; \;  \forall t \in [0,T].
\end{equation}

Equation (\ref{eqn_19}) can be rearranged by taking logs on both sides as:

\begin{equation}\label{log_eqn}
     \ln  \frac{\Vert \xi(\zeta_0,t) - \xi(\zeta'_0,t)\Vert}{\Vert \zeta_0 - \zeta'_0 \Vert}  \leq \ln K + {\hat{\gamma} t},\; \;  \forall t \in [0,T].  
\end{equation}

 The above inequality has a general structure of: 
 \vspace{-0.5cm}
\begin{equation}\label{eq:18}
     \mu \leq a\nu + b,\; \; \forall (\mu,\nu) \in \Gamma.
\end{equation}
where for $\Gamma \subseteq \mathbb{R} \times \mathbb{R}$, a pair $(a,b)$ is a linear separator and $(\mu,\nu)$ represents $\left( \ln\frac{\Vert \xi(\zeta_0,t) - \xi(\zeta'_0,t)\Vert}{\Vert \zeta_0 - \zeta'_0 \Vert},t \right)$ in (\ref{log_eqn}). Therefore, the learning task is identifying the $(a,b)$ values from sampling points that make the inequality in (\ref{eq:18}) a linear separator for the large portion of points in $\Gamma$. The sampling points are assumed to be drawn based on unknown distribution $\mathcal{D}$. The probabilistic algorithm provided in Algorithm \ref{reachability_analysis} has been proposed in \cite{10.1007/978-3-319-63387-9_22} to identify the appropriate values of $(a,b)$. The separator discovered by the above algorithm has a correctness guarantee with high probability. The proof can be obtained in \cite{10.1007/978-3-319-63387-9_22}. To minimize the conservative nature of the discrepancy function, we adopt a piece-wise exponential discrepancy function of the form $\beta(\zeta_0,\zeta_0',t) =  \Vert \zeta_0 - \zeta'_0 \Vert Ke^{\sum_{j=1}^{i-1}\gamma_j(t_j - t_{j-1}) + \gamma_i(t - t_{i-1}) }$ from \cite{10.1007/978-3-319-63387-9_22}. This enables us to divide the time window for the reachable set into several smaller segments and find discrepancy parameters for each segment, resulting in less conservative reachable bounds.

\begin{figure}[!ht]
    \centering 
\begin{subfigure}{0.5\textwidth}
  \includegraphics[width=\linewidth]{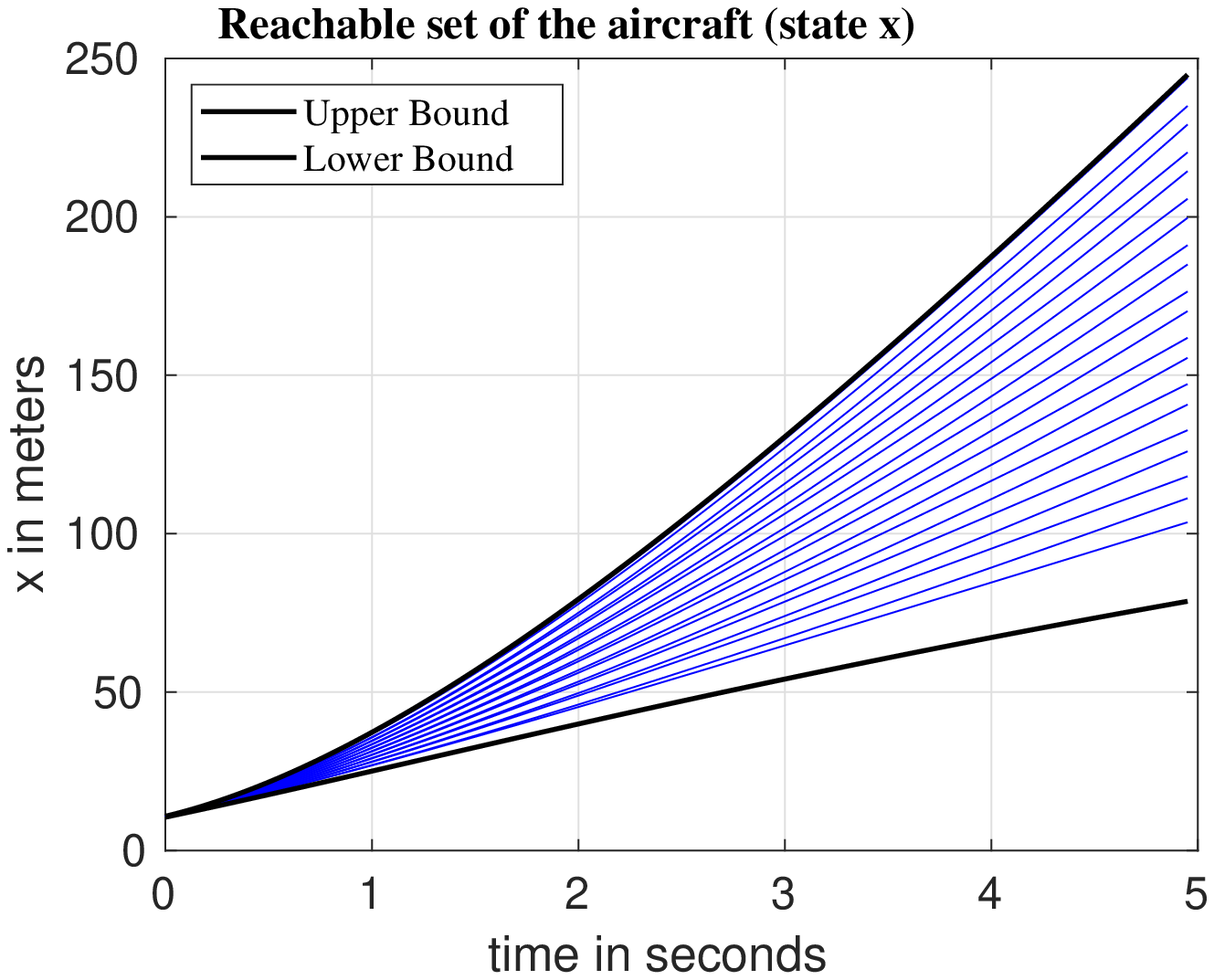}
  \caption{Reachable set of the aircraft (state x)}
  \label{fig:state_x}
\end{subfigure}\hfil 
\begin{subfigure}{0.5\textwidth}
  \includegraphics[width=\linewidth]{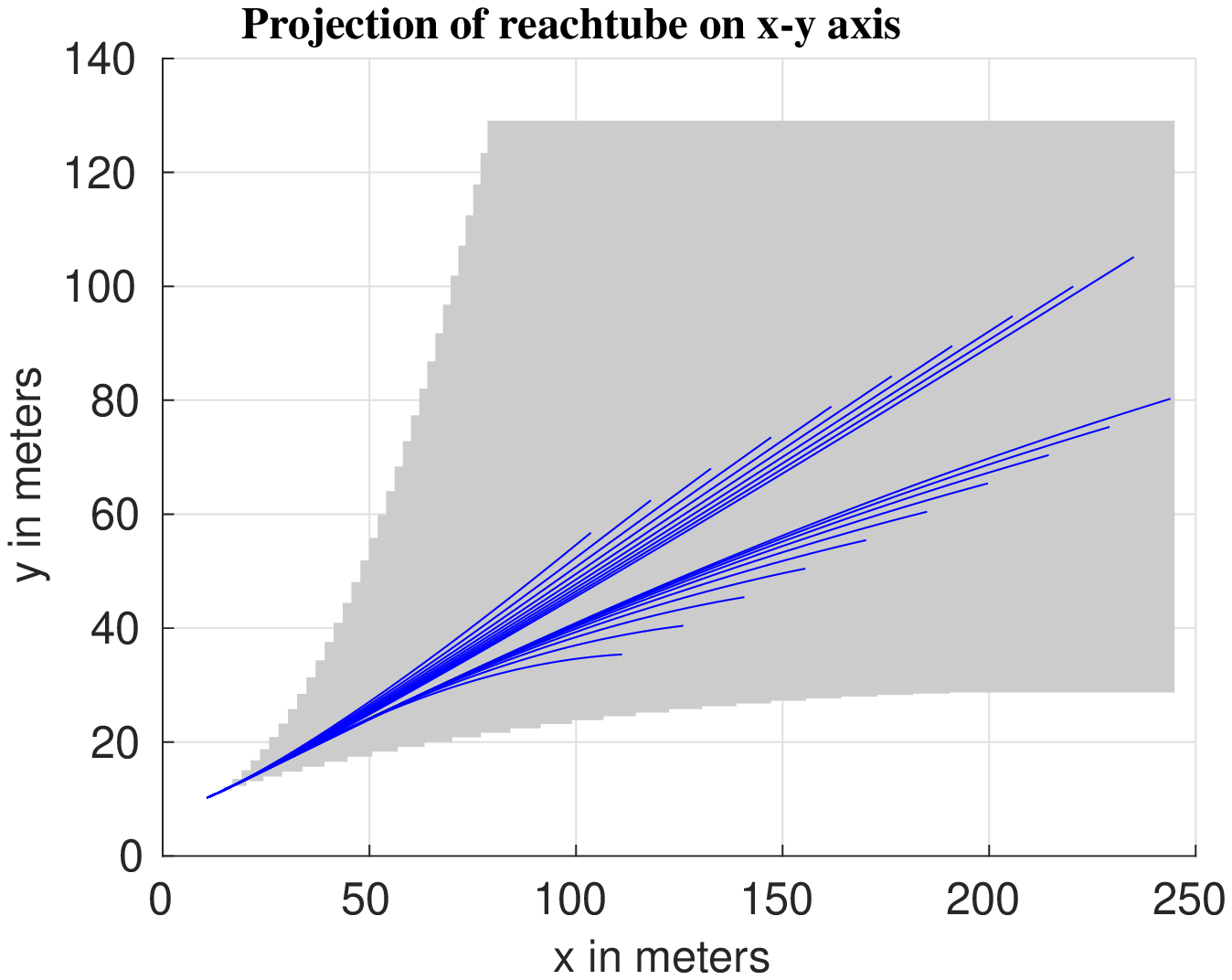}
  \caption{Projection of reach-tube on x-y axis}
  \label{fig:state_xy}
\end{subfigure}\hfil 

\medskip

\begin{subfigure}{0.5\textwidth}
  \includegraphics[width=\linewidth]{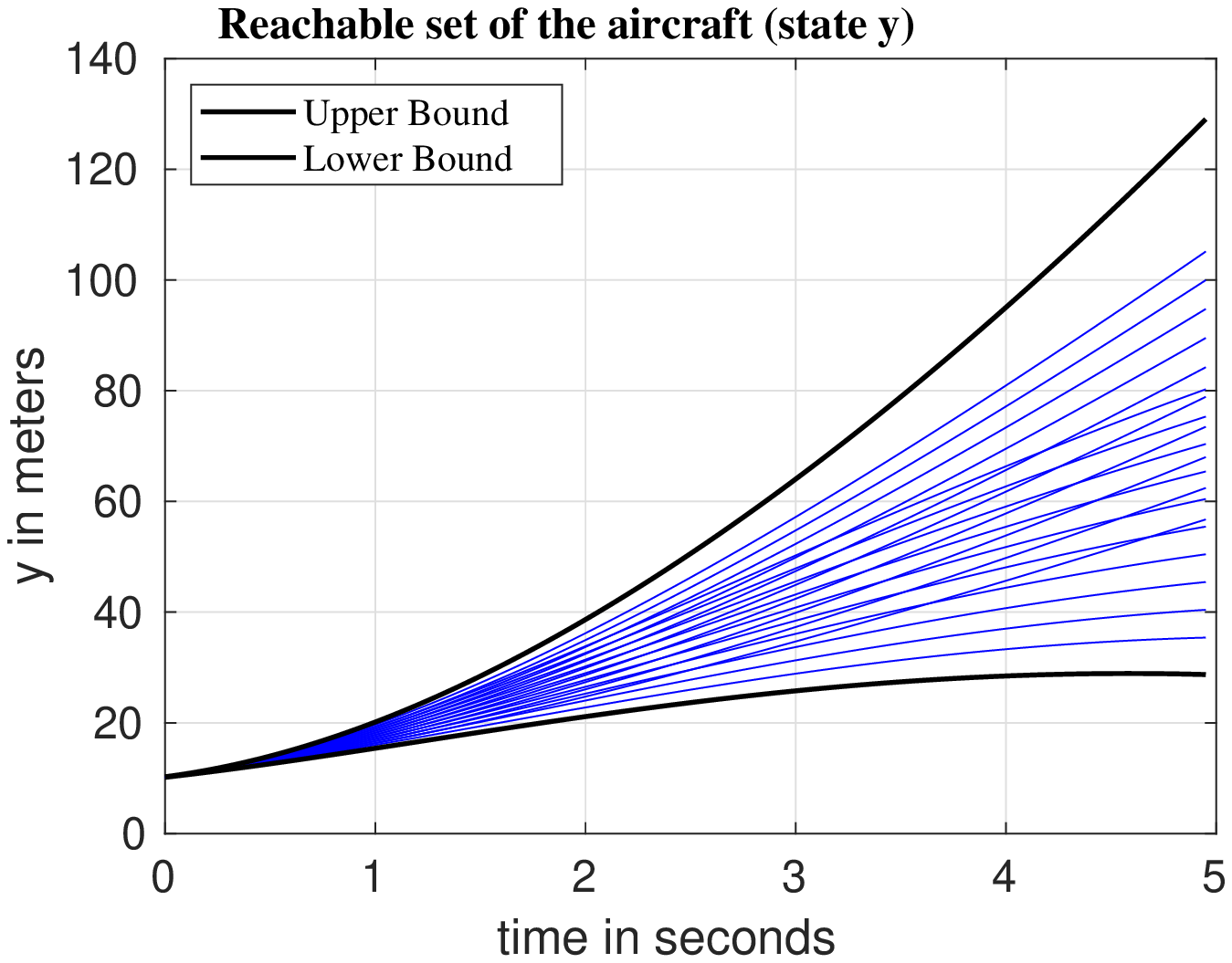}
  \caption{Reachable set of the aircraft (state y)}
  \label{fig:state_y}
\end{subfigure}\hfil 
\begin{subfigure}{0.5\textwidth}
  \includegraphics[width=\linewidth]{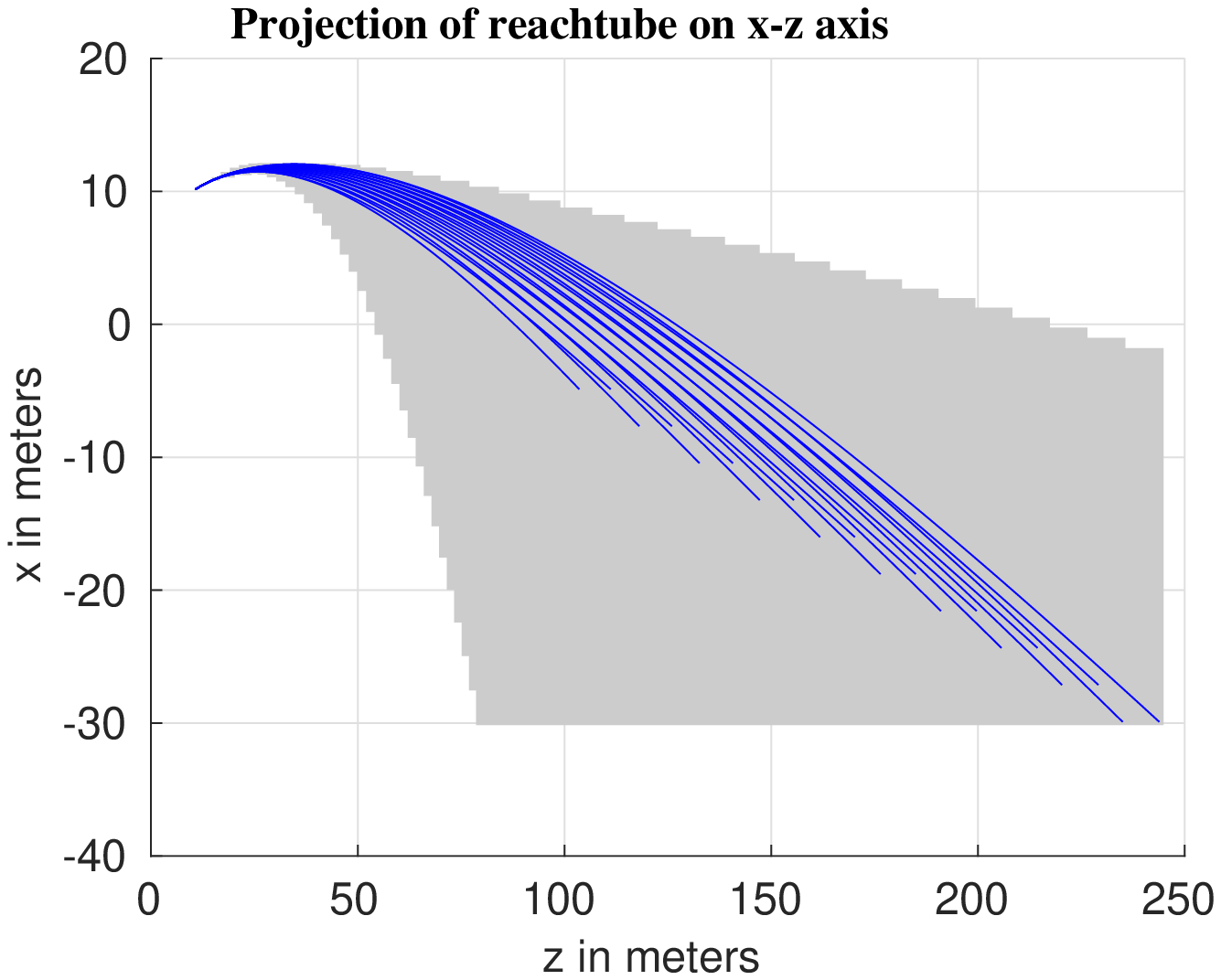}
  \caption{Projection of reach-tube on x-z axis}
  \label{fig:state_xz}
\end{subfigure}\hfil 

\medskip
\begin{subfigure}{0.5\textwidth}
  \includegraphics[width=\linewidth]{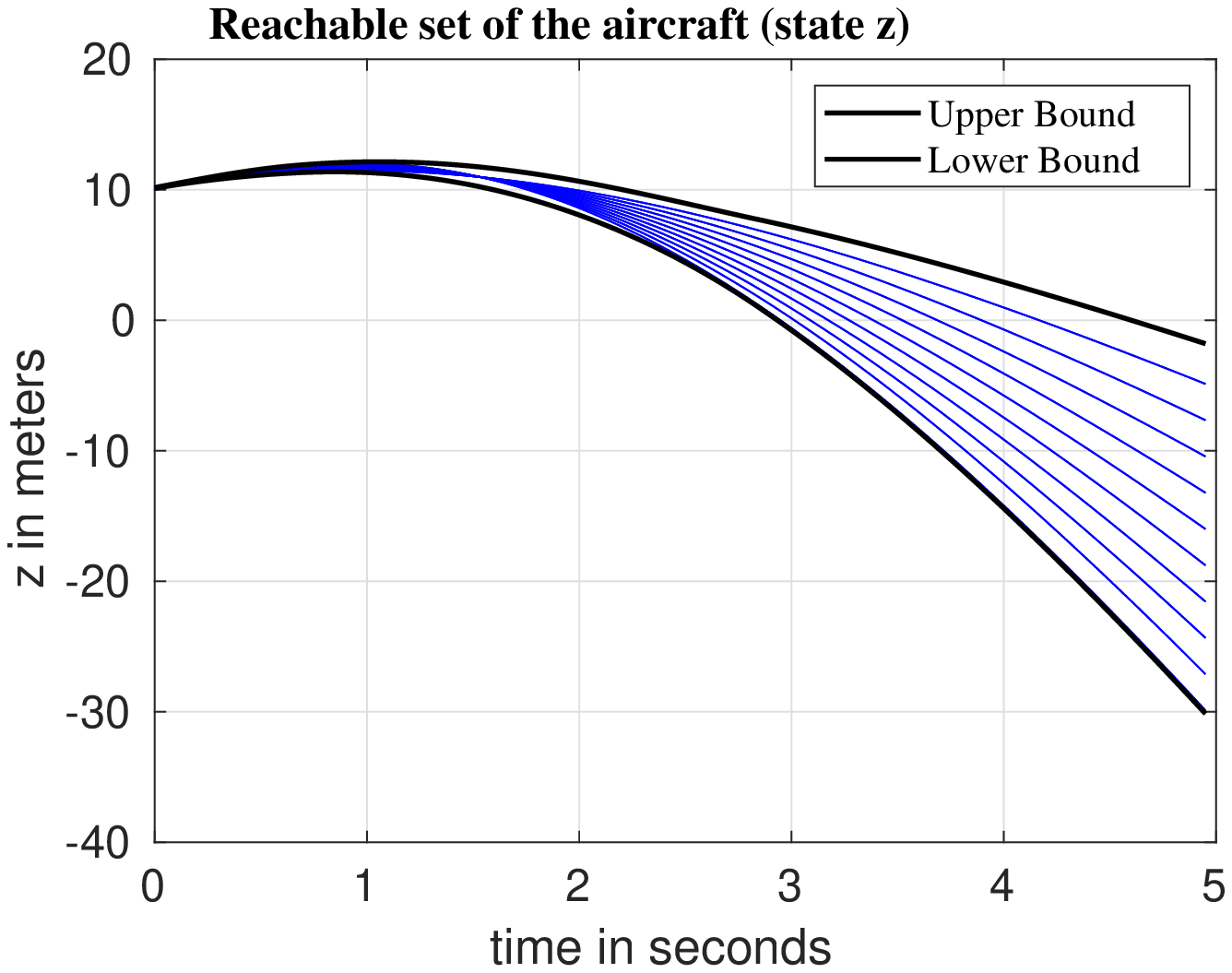}
  \caption{Reachable set of the aircraft (state z)}
  \label{fig:state_z}
\end{subfigure}\hfil 
\begin{subfigure}{0.5\textwidth}
  \includegraphics[width=\linewidth]{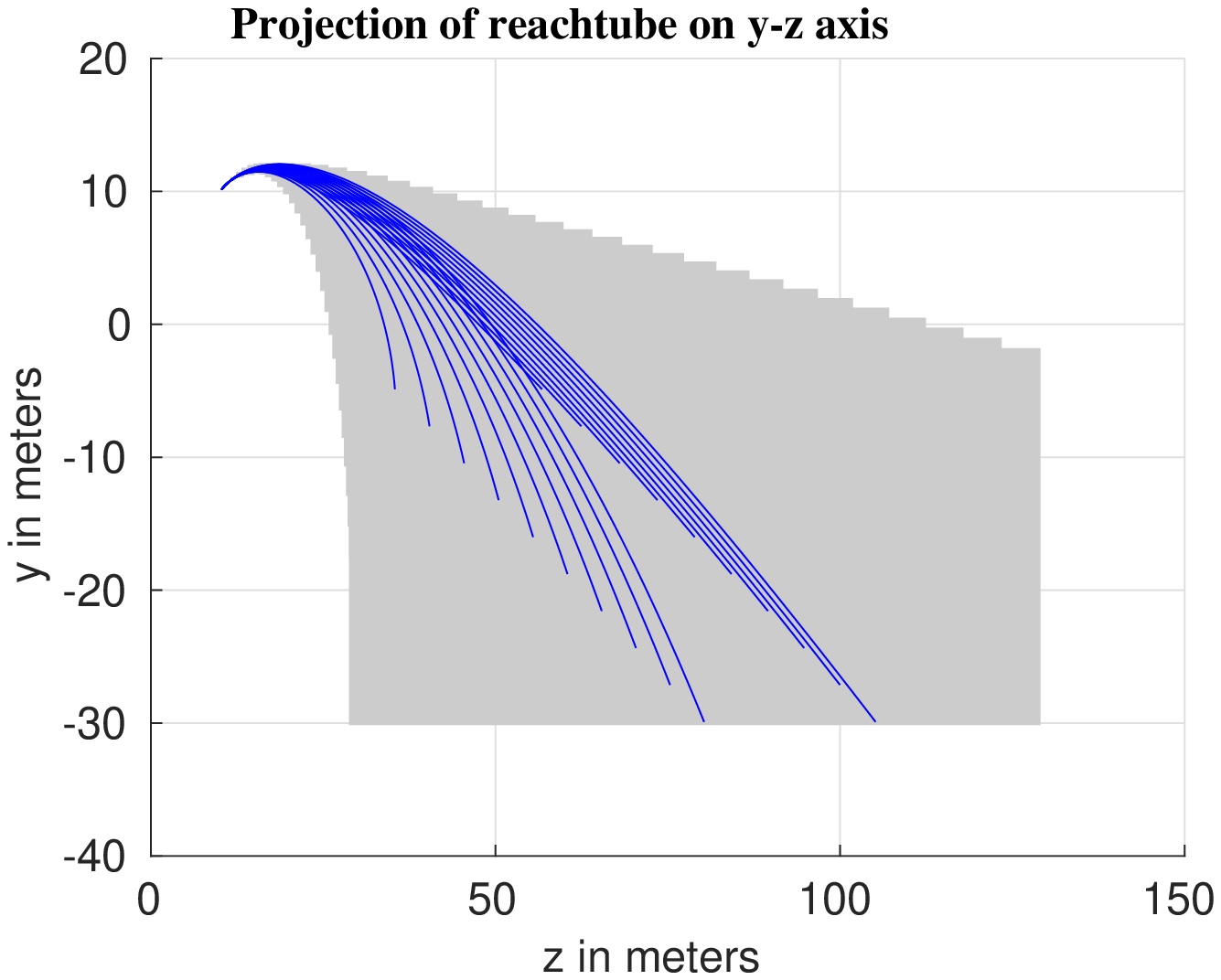}
  \caption{Projection of reach-tube on y-z axis}
  \label{fig:state_yz}
\end{subfigure}
\caption{Generated reachable set using Algorithm \ref{reachability_analysis}.}
\label{fig:reach_set}
\end{figure}

\FloatBarrier

The procedure to over-approximate the reachable set of an aircraft is outlined in Algorithm \ref{reachability_analysis}. The inputs to the algorithm include the aircraft dynamics, the action set $\mathcal{A}$, the initial states of the aircraft $\zeta_0$, and the time horizon $T$. The algorithm then generates trajectories by randomly choosing from the set of control actions.  It then computes the maximum pair-wise distance between the initial states and each trajectory using Chebyshev distance and gets the sensitivity parameters for each time step ($\mu(t)$ and $\nu(t)$). The convex hull of these parameters is then determined, and the values $a$ and $b$ are obtained, which represent the discrepancy function $K$ and $\hat{\gamma}$.


\begin{algorithm}
\SetKwProg{Fn}{Procedure}{\string:}{}
\SetKwFunction{FRecurs}{ReachabilityAnalysis}%
\Fn(){\FRecurs{}}{ 
\textbf{Input : }Action set $\mathcal{A}$, aircraft dynamics $\dot{\zeta}(t)$, initial state $\zeta_0$, time horizon $T$ 

\textbf{Output : }Reachable set $\mathcal{R}_i(\zeta_i(t),T)$

\nl $\Gamma(t) \leftarrow f(\zeta(t), \mathcal{A})$\tcc*[r]{randomly sample from $\mathcal{A}$ and generate a set of trajectories}

\nl $\Vert \zeta_0 - \zeta'_0 \Vert \leftarrow \mathcal{D}_{\mathbb{C}}(\Gamma(t_0))$ \tcc*[r]{compute distance between initial states}

\nl $\Vert \xi(\zeta_0,t) - \xi(\zeta'_0,t) \Vert \leftarrow \mathcal{D}_{\mathbb{C}}(\Gamma(t))$\tcc*[r]{compute distance between trajectories}

\nl $\mu(t) \leftarrow \ln\frac{\Vert \xi(\zeta_0,t) - \xi(\zeta'_0,t)\Vert}{\Vert \zeta_0 - \zeta'_0 \Vert} $\tcc*[r]{compute sensitivity parameters}

\nl $\nu(t) \leftarrow t$

\nl $ \sum_i^n \mu_i = \nu_i a_i + b_i \leftarrow \mathtt{covhull}(\mu(t),\nu(t))$\tcc*[r]{compute discrepancy parameters}

\nl $a_i \leftarrow \frac{\Delta \mu_i}{\Delta t}$, $b_i \leftarrow \mu_i - \nu_i a_i$

\nl $\beta(\zeta_0,\zeta_0',t) \leftarrow  \Vert \zeta_0 - \zeta'_0 \Vert Ke^{\sum_{j=1}^{i-1}\gamma_j(t_j - t_{j-1}) + \gamma_i(t - t_{i-1}) }$\tcc*[r]{compute the piece-wise exponential discrepancy function}

\nl $\mathcal{R}_i(\zeta_i(t),T) \leftarrow \beta(\zeta_0,\zeta_0',t)$
}
\caption{Reachability Analysis \label{reachability_analysis}}
\end{algorithm}

Figures \ref{fig:state_x} to \ref{fig:state_yz} show how a reachable set of aircraft can be over-approximated by simulating several trajectories from the current state. Figures \ref{fig:state_x}, \ref{fig:state_y}, and \ref{fig:state_z} depict the reachable sets of $x$, $y$, and $z$ states of the aircraft, respectively. Figure \ref{fig:state_xy}, \ref{fig:state_xz}, and \ref{fig:state_yz} show the projections of the reach-tube of an aircraft on different planes.

\begin{figure}[!ht]
\begin{center}
\includegraphics[width=1.05\textwidth]{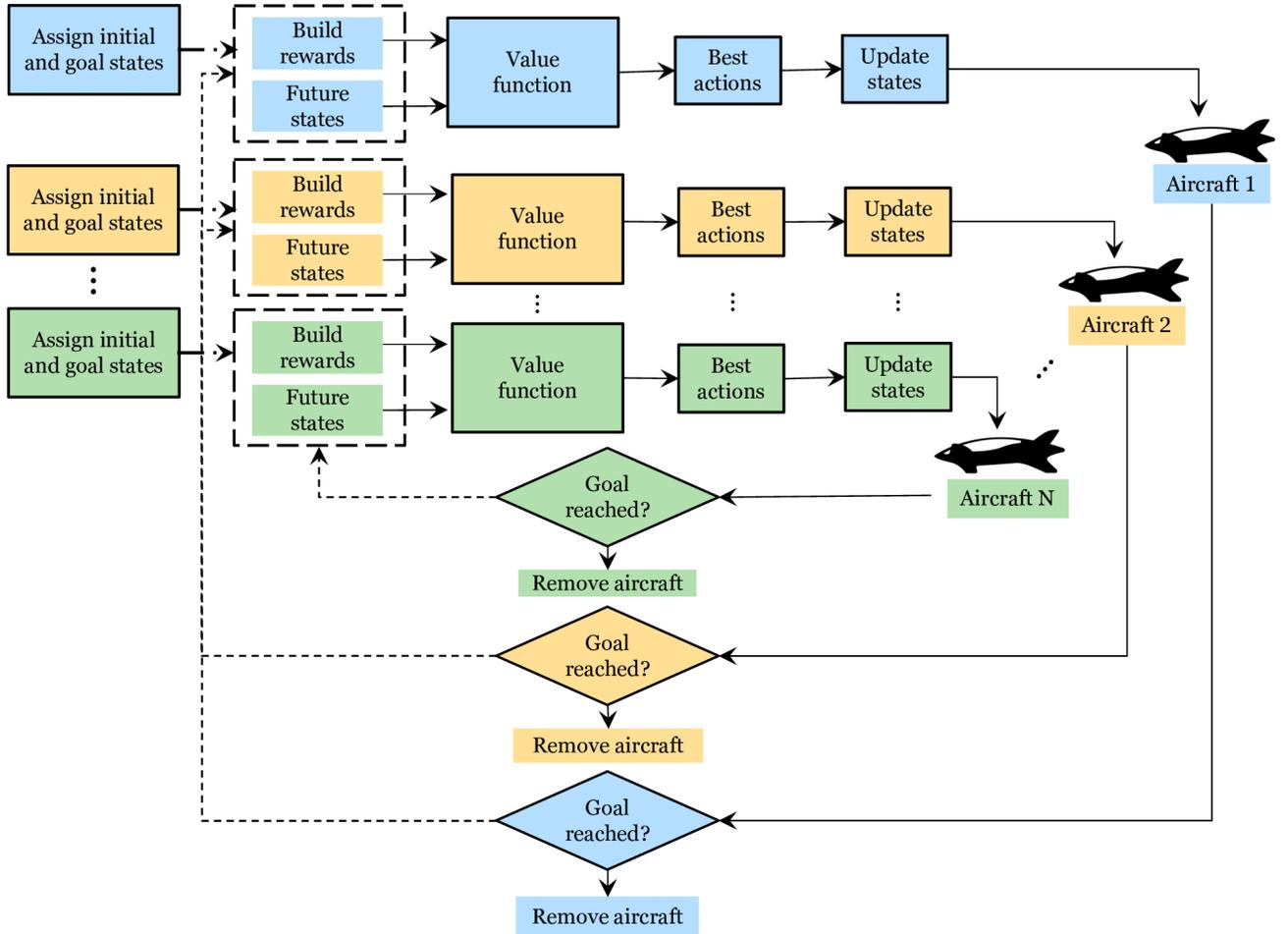}
\caption{\textbf{Schematic diagram representation of the trajectory planner. The framework assigns initial and goal states for each aircraft and identifies reward sources. It, then, projects future states and computes their values with the value function. Finally, the best action yielding the maximum reward will be selected and the aircraft states will be updated accordingly. This process is repeated iteratively until each aircraft reaches its destination vertiport.}}
\label{fig:schematic_diagram}
\end{center}
\end{figure}

Figure \ref{fig:schematic_diagram} illustrates the overall operational procedure of the proposed trajectory planner. As shown in the figure, the framework first assigns initial and goal states for each aircraft in the system. Subsequently, for each aircraft, it identifies the positive and negative reward sources as discussed in \ref{ssec:reward}. After the reward sources are identified, it forward projects the future states of the aircraft using the action sets and computes the values of each future state using the value function as given in Equation \ref{value_function}. The best action yielding the maximum total reward is then selected, and the states of the aircraft are updated using the chosen control action. This process is repeated iteratively for each aircraft until each aircraft reaches its designated destination vertiport.

\subsection{The Proposed Trajectory Planning Framework}
\vspace{0.1cm}
 The detailed working procedure of the trajectory planning is provided in Algorithm \ref{pseudocode} and the schematic diagram in Figure \ref{fig:schematic_diagram}. Here, we highlight the two main modules: Reachability Analysis and Trajectory Planner.   

\textit{Trajectory Planner:} The proposed framework works in a decentralized manner, where each aircraft will be responsible for choosing a control action that satisfies the reach-avoid property defined below. To achieve this, it first forward projects the future states of an aircraft using the dynamics of the aircraft and the control actions provided in the action space. Then, it computes the positive and negative rewards for the projected states and picks the control action that maximizes the total reward. 

\textit{Reachability Analysis:} While building the negative rewards, the framework considers the reachable sets of nearby intruder aircraft and the terrain around the aircraft. The algorithms discussed in section \ref{Rechability_subsection} will be utilized to compute the reachable sets.

\begin{algorithm}[h!]
\SetAlgoNoEnd
\SetKwProg{Fn}{Procedure}{\string:}{}
\SetKwFunction{FRecurs}{TrajectoryPlanner}%
\Fn(){\FRecurs{world state}}{
\nl $S_0 \leftarrow $ randomly initialize aircraft states

\nl \Repeat{ \textnormal{each aircraft reaches its final destination} }{
\nl \For{each aircraft i }{
\nl $\zeta_t \leftarrow$ current state of the ownship

\nl $\Gamma(t) \leftarrow f(\zeta(t), \mathcal{A})$ \tcc*[r]{project future states of the ownship using the action set}

\nl $P^+ \leftarrow $ vertiport location \tcc*[r]{build positive reward for destination}

\nl $\zeta_j \leftarrow$ identify nearby aircraft

\nl $\mathcal{R}_i(\zeta_i(t),T) \leftarrow$ \textbf{\tt Reachability Analysis$(\zeta_j)$} \tcc*[r]{compute the reach set using Algorithm \ref{reachability_analysis}} 

\nl $P^- \leftarrow \mathcal{R}_i(\zeta_i(t),T)$  \tcc*[r]{build negative reward} 
\nl \For{$\zeta \in \Gamma$ }{ 
\nl $d_p \leftarrow \Vert \zeta_j - \textbf{\text{\tt location}}(P^+) \Vert_{2}$ 

\nl $r_p \leftarrow \textbf{\text{\tt reward}}(P^+)$

\nl $\gamma_p \leftarrow \textbf{\text{\tt discount}}(P^+)$

\nl $V^+_{p} \leftarrow |r_p|\cdot \gamma_p^{d_p}$ \tcc*[r]{compute positive values for each future state}

\nl $\textbf{V}_{\text{\tt max}}^+ \leftarrow \underset{p}{\mathrm{max}} \textbf{V}_{p}^+$ 

}
\nl \For{$n_i \in P^-$}{
\nl $d_n \leftarrow \Vert\zeta_j - \textbf{\text{\tt location}}(n_i) \Vert_{2}$ 

\nl $\rho_n \leftarrow d_n < \textbf{\text{\tt radius}}(n_i) $ 

\nl $r_n \leftarrow  \textbf{\text{\tt reward}}(n_i) $ 

\nl $\gamma_n \leftarrow  \textbf{\text{\tt discount}}(n_i) $

\nl $V^-_{n_i} \leftarrow \text{\tt int}(\rho_n) \cdot |r_n|\cdot \gamma_n^{d_n}$ \tcc*[r]{compute negative values for each future state}

}
\nl \If{ $\text{\tt altitude}(\zeta_t) < \text{\tt penalty altitude}$}{
\nl $V_{\text{\tt terrain}} \leftarrow 1000 - \text{\tt altitude}(\zeta_t)$

\nl \Else{$V_{\text{\tt terrain}} \leftarrow 0$ 

}
}
\nl $\textbf{V}^*[\zeta_i] \leftarrow V_{\text{\tt max}}^+ - V_{\text{\tt max}}^- - V_{\text{\tt terrain}}$ \tcc*[r]{compute total values for each future state}

\nl $i_{\text{\tt max}} \leftarrow \underset{\zeta}{\mathrm{argmax}}(\textbf{V}^*)$

\nl $\zeta_{t+1} \leftarrow Z_1[i_{\text{\tt max}}]$

\nl  $S_{t+1}[i] \leftarrow \zeta_{t+1}$

}
}
}
\caption{ Online Verified Trajectory Planning Framework \label{pseudocode}}
\end{algorithm}

\textit{Reach-avoid property}: For an aircraft starting from an initial state $\zeta(0)$, we say the reach-avoid property is satisfied if and only if its trajectory $\zeta(t)$, (1) never enters into an unsafe set $\mathcal{S}_u$, and (2) reaches a goal set $\mathcal{S}_g$ within a finite time horizon $T$. These two conditions can be expressed mathematically as follows:   

\begin{equation}\label{reach_avoid}
    ( \forall t \in 0\leq t \leq T, \; \xi(\zeta(0),t) \cap \mathcal{S}_u = \emptyset) \bigwedge (\exists \; t \; 0\leq t\leq T, \; \xi(\zeta(0),t) \cap \mathcal{S}_g \neq \emptyset)
\end{equation}   
In the above equation, the unsafe set $\mathcal{S}_u$ is composed of the reachable sets of nearby intruders and the terrain.

\textbf{Theorem 1:} Consider aircraft $i$ has access to other nearby intruder aircraft's dynamics and current states. In addition, consider aircraft $i$ has information about the environment's terrain. Then, aircraft $i$ can choose a control action from the action space $\mathcal{A}$ for its next state that is guaranteed to satisfy the reach-avoid property given in Equation \ref{reach_avoid}. 

\textit{Proof:} Consider the reach-avoid property is not satisfied for aircraft $i$. Such an assumption entails that either the aircraft has entered an unsafe state $\mathcal{S}_u$, or it is not progressing to its goal state $\mathcal{S}_g$. However, because the reachable sets of nearby aircraft and the terrain information are accessible, it can choose a control action that enables the aircraft to avoid entering the reachable sets of nearby aircraft. In addition, since the MDP-based trajectory planner generates a reward that motivates the aircraft to move to its destination, aircraft $i$ will always progress towards its destination. Hence, Theorem 1 is true by contradiction.   \QEDA 
\section{Results and Discussion}
\label{sec:results}

In this section, the performance of the proposed method is discussed. Since the objective of this paper is to develop a safe and scalable UAM trajectory planning framework, the two criteria we used to evaluate the performance of the proposed algorithm are mean computational time and the number of Near Mid Air Collisions (NMAC). Mean computation time, which is the time taken in each step by the algorithm to compute the safe trajectory for a single aircraft, demonstrates the computational efficiency of the method. On the other hand, NMAC, defined as a loss of $152$ meters of horizontal and $30$ meters of vertical separation \cite{chen2023integrated}, is used to evaluate the ability of the algorithm to guide the aircraft and avoid collisions.

\subsection{Scenario Description}
\begin{figure}[h!]
\begin{center}
\includegraphics[width=0.9\textwidth]{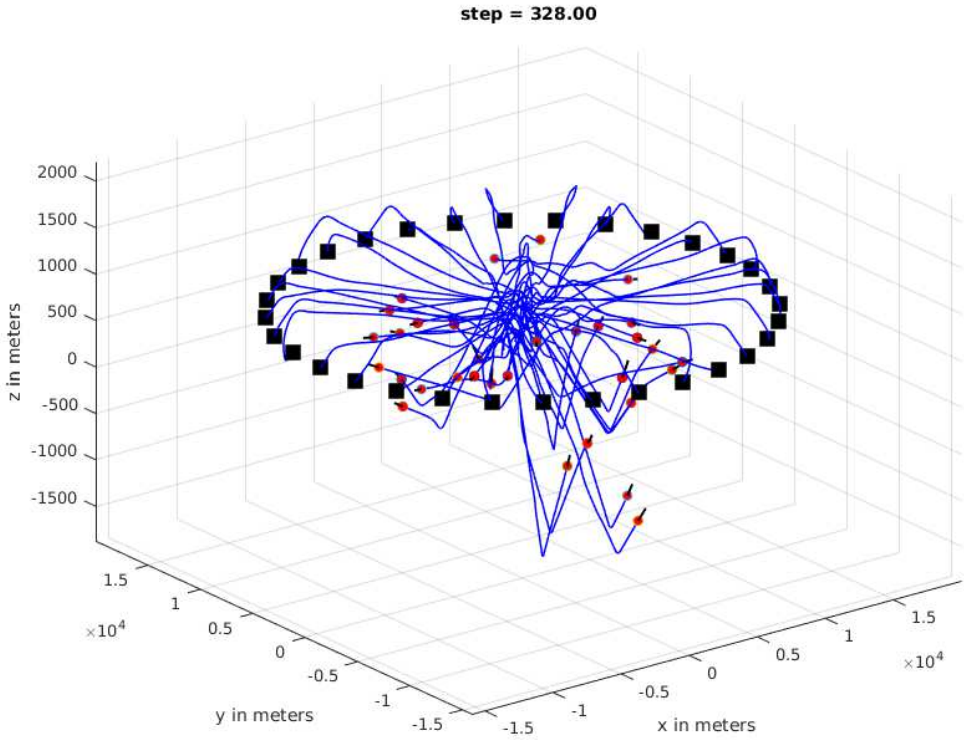}
\caption{\textbf{Snapshot of the simulation environment. The environment simulates the operation of 32 aircraft, each rendered as a red circle. The black boxes represent the vertiports where the aircraft take off and navigate towards. The black lines represent the aircraft trajectories for the next ten steps. The blue lines indicate the paths traveled by aircraft.}}
\label{fig:simulation}
\end{center}
\end{figure}

A snapshot of the simulation environment we used to evaluate the performance of the  trajectory planning framework is shown in Fig \ref{fig:simulation}. The simulation defined a geographical bounding box that encompasses a volume of $15km$ radius. The aircraft are assigned to take off from their origin vertiports and fly to destination vertiports located on the opposite side of their origin. The environment is configurable to accommodate a variable number of vertiports and aircraft, which utilize the proposed trajectory planning framework in a distributed manner. 

In reference to the designed scenario, it is important to note that, as depicted in Figure \ref{fig:simulation}, all aircraft are scheduled to travel through a central location in the environment. This scenario, although unlikely to occur in a typical UAM setting, serves as a means to evaluate the detect-and-avoid (DAA) capabilities of the system under adverse conditions where strategic deconfliction fails.

We present experimental results on a different number of aircraft assigned to fly to their designated goal states. The algorithm utilized in these experiments has been implemented using MATLAB. Additionally, a video demonstration showcasing the results of the algorithm for 8\footnote{https://youtu.be/9ycsue5bhb4}, 16\footnote{https://youtu.be/inyiLlfCNns}, and 32\footnote{https://youtu.be/iqxr-0Zkh3Q} aircraft can be viewed on YouTube. 

All experiments were conducted on a 3.20 GHZ Intel Xeon (R) CPU with 125.4 GB RAM. Each experiment was repeated 25 times for each aircraft number, with randomly generated initial locations for the aircraft. The computational time and NMACs for each aircraft number are reported.

\begin{table}[!htb]
\begin{minipage}{.5\linewidth}
\caption{NMAC performance }
\centering
\begin{tabular}{|| c c c ||}
\hline 
 Aircraft & mean & std \\ [0.5ex]
 \hline \hline
 $2$ & $0$ & $0$  \\
\hline
 $4$ & $0$ & $0$  \\
\hline
$8$ & $0$ & $0$  \\
\hline
$16$ & $1.36$ & $5.12$  \\
\hline
$32$ & $9.96$ & $10.91$  \\
\hline
\end{tabular}\label{NMAC_1}
\end{minipage}%
\begin{minipage}{.5\linewidth}
\centering
\caption{Computation time performance}
\begin{tabular}{|| c c c c||}
\hline 
 Aircraft & mean (sec) & std (sec)& throughput (sec)\\ [0.5ex]
 \hline \hline
 $2$ & $0.03$ & $0.03$& $31.05$  \\
\hline
 $4$ & $0.05$ & $0.06$ & $99.46$ \\
\hline
$8$ & $0.11$ & $0.14$ & $489.16$ \\
\hline
$16$ & $0.17$ & $0.21$ & $1815.25$ \\
\hline
$32$ & $0.27$ & $0.29$ & $6448.43$ \\
\hline
\end{tabular}\label{time_1}
\end{minipage} 
\end{table}

The experimental results demonstrate the effectiveness of the proposed trajectory planner in guiding the motion of each aircraft from its initial position to its assigned vertiport. Tables \ref{NMAC_1} and \ref{time_1} present the trajectory planner's NMAC and computational time performances of the trajectory planner. As shown in Table \ref{time_1}, the mean computational time of the framework increases as the number of aircraft in the system increases, but it grows in a polynomial order with the increased number of aircraft, indicating the scalability of the approach. Table \ref{time_1} also presents the throughput performance of the algorithm, defined as the amount of time taken to guide each aircraft in the system to its assigned vertiport successfully.

On the other hand, despite utilizing a formal verification scheme based on reachability analysis, as indicated in Table \ref{NMAC_1}, there were instances of NMACs observed in the environment as the number of aircraft increased. This is primarily due to the fact that the MDP formulation converts hard constraints, such as collisions, into benign conditions represented by negative rewards. As a result, in congested environments, there may be instances of momentary violations of safety constraints. In the subsequent sections, we will discuss the methods employed to address this issue.

\subsection{Action Shielding}
\begin{figure}[!ht]
    \centering 
  \includegraphics[width=.95\linewidth]{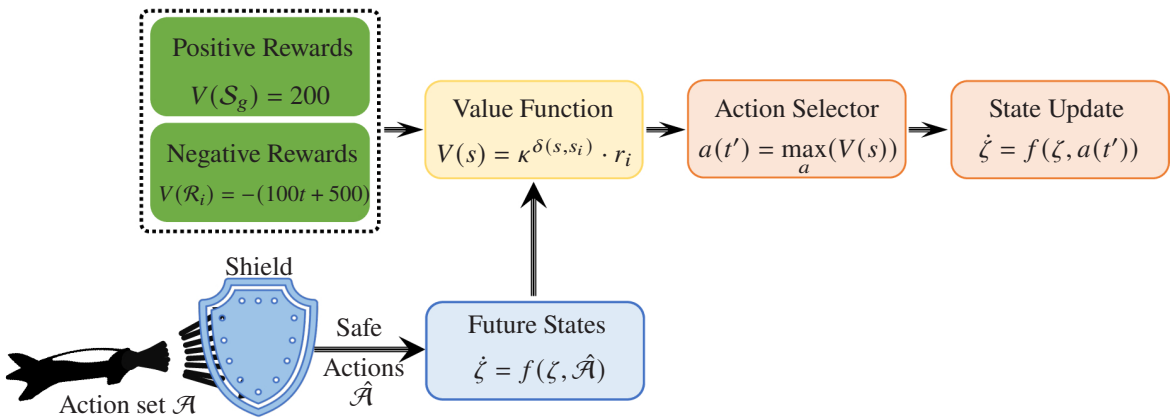}
  \put(-420,5){Action set $\mathcal{A}$}
  \put(-357,57){Shield}
  \put(-315,34){Safe}
  \put(-320,19){Actions}
  \put(-310,8){$\mathcal{\hat{A}}$}
  \put(-263,18){$\dot{\zeta} = f(\zeta,\mathcal{\hat{A}})$}
  \put(-265,35){Future States}
  \put(-271,117){Value Function}
  \put(-277,100){$V(s) = \kappa^{\delta(s,s_i)} \cdot r_i$}
  \put(-172,117){Action Selector}
  \put(-178,102){$a(t') = \underset{a}{\mathrm{max}}(V(s))$}
  \put(-68,117){State Update}
  \put(-72,102){$\dot{\zeta} = f(\zeta,a(t'))$}
  \put(-377,140){Positive Rewards}
  \put(-370,122){$V(\mathcal{S}_g) = 200$}
    \put(-379,100){Negative Rewards}
  \put(-382,85){\scalebox{0.85}{$ V(\mathcal{R}_i) = -(100t + 500)$}}
  \caption{Action shielding procedure. The implemented shield filters aircraft actions that result in unsafe states. After obtaining the set of safe actions, the value function is employed to calculate the values of future states, considering the positive and negative rewards within the environment. Subsequently, the action selector chooses the action that yields the maximum total reward, and the aircraft's states are updated based on the selected action.}
  \label{fig:Action_shield}
\end{figure}

One potential solution to the challenge of enforcing hard constraints on an MDP agent is through the implementation of action shielding \cite{konighofer2020shield}, in which the agent's actions are filtered through a mechanism that blocks actions that result in unsafe states, as shown in Figure \ref{fig:Action_shield}. The value of states is utilized to filter out actions that lead to unsafe states. Specifically, if the value of a state resulting from a certain action is negative, the shield will eliminate the action from the set of valid actions. However, in instances where all control actions lead to unsafe states, this technique results in a deadlock as all actions are blocked. To circumvent this scenario, we propose an alternative control action for a short time horizon. It is worth noting that this approach may result in violations of state constraints imposed for passenger comfort, as safety is given priority over comfort. As such, the new control action set (in degree) to be implemented during a deadlock will be: 

\begin{equation}
\gamma^c = [-180, -139.5,  -99,  -58.5,  -18,   18,   58.5,   99,  139.5, 180]
\end{equation}
\begin{equation}
\phi^c = [-180, -139.5,  -99,  -58.5,  -18,   18,   58.5,   99,  139.5, 180]
\end{equation}


Tables \ref{NMAC_2} and \ref{time_2} present the performance of the trajectory planner with the added enhancement of action shielding with regard to the number of NMACs and computational time, respectively. As shown in Table \ref{NMAC_2}, it is evident that the addition of action shielding has resulted in a significant improvement in the safety performance of the trajectory planner. However, as demonstrated in Table \ref{time_2}, the change in the computational time is minimal. 

\begin{table}[!htb]
\begin{minipage}{.5\linewidth}
\caption{NMAC performance }
\centering
\begin{tabular}{|| c c c ||}
\hline 
 Aircraft & mean & std \\ [0.5ex]
 \hline \hline
 $2$ & $0$ & $0$  \\
\hline
 $4$ & $0$ & $0$  \\
\hline
$8$ & $0$ & $0$  \\
\hline
$16$ & $0.68$ & $2.35$  \\
\hline
$32$ & $2.12$ & $5.13$  \\
\hline
\end{tabular}\label{NMAC_2}
\end{minipage}%
\begin{minipage}{.5\linewidth}
\centering
\caption{Computation time performance}
\begin{tabular}{|| c c c c||}
\hline 
Aircraft & mean (sec) & std (sec)& throughput (sec)\\ [0.5ex]
 \hline \hline
 $2$ & $0.03$ & $0.02$& $30.00$  \\
\hline
 $4$ & $0.05$ & $0.06$ & $99.57$ \\
\hline
$8$ & $0.11$ & $0.13$ & $467.07$ \\
\hline
$16$ & $0.16$ & $0.20$ & $1724.92$ \\
\hline
$32$ & $0.22$ & $0.27$ & $6174.32$ \\
\hline
\end{tabular}\label{time_2}
\end{minipage} 
\end{table}

\subsection{Reward Shaping}
Many existing techniques in the literature address the issue of undesirable behavior exhibited by MDP agents through the use of reward engineering or reward shaping. Reward shaping refers to the process of modifying the reward received by the agent to elicit desired behavior, as outlined in \cite{memarian2021self}. In other words, instead of using the traditional MDP $M = (S,A,T,\kappa, R)$, we use a transformed MDP $M' = (S, A, T,\kappa, R')$, where $R' = R+F$ is the reward function in the transformed MDP, and $F: S \times A \times S \rightarrow R$ is a bounded real-valued function known as the reward-shaping  function. The specific reward shaping function employed in this study is a difference of potentials $F(s,a,s') = \Phi(s') - \Phi(s)$, where $\Phi$ is the value function over states \cite{ng1999policy}. 
\begin{equation}
    F(s,a,s') = \kappa(V^*(s')) - V^*(s),
\end{equation}
where, $\kappa$ is the discount factor and $V^*(s')$ and $V^*(s)$ are the values of the current and future states.

Tables \ref{time_3} and \ref{NMAC_3} present the performance of the trajectory planner with the enhancement of reward shaping in terms of the number of NMAC and computational time, respectively. From Table \ref{NMAC_3}, we can see that the reward shaping technique has led to a superior improvement in safety performance when compared to the action shielding technique. However, the impact on computational time is negligible.

\begin{table}[!htb]
\begin{minipage}{.5\linewidth}
\caption{NMAC performance }
\centering
\begin{tabular}{|| c c c ||}
\hline 
 Aircraft & mean & std \\ [0.5ex]
 \hline \hline
 $2$ & $0$ & $0$  \\
\hline
 $4$ & $0$ & $0$  \\
\hline
$8$ & $0$ & $0$  \\
\hline
$16$ & $0.24$ & $1.20$  \\
\hline
$32$ & $1.48$ & $3.63$  \\
\hline
\end{tabular}\label{time_3}
\end{minipage}%
\begin{minipage}{.5\linewidth}
\centering
\caption{Computation time performance}
\begin{tabular}{|| c c c c||}
\hline 
 Aircraft & mean (sec) & std (sec)& throughput (sec)\\ [0.5ex]
 \hline \hline
 $2$ & $0.03$ & $0.02$& $29.68$  \\
\hline
 $4$ & $0.05$ & $0.06$& $101.37$  \\
\hline
$8$ & $0.11$ & $0.13$& $463.09$  \\
\hline
$16$ & $0.18$ & $0.23$& $1968.56$  \\
\hline
$32$ & $0.23$ & $0.27$ & $6311.54$ \\
\hline
\end{tabular}\label{NMAC_3}
\end{minipage} 
\end{table}

\begin{figure}[h]
    \centering 
\begin{subfigure}[b]{0.45\textwidth}
  \includegraphics[width=\linewidth]{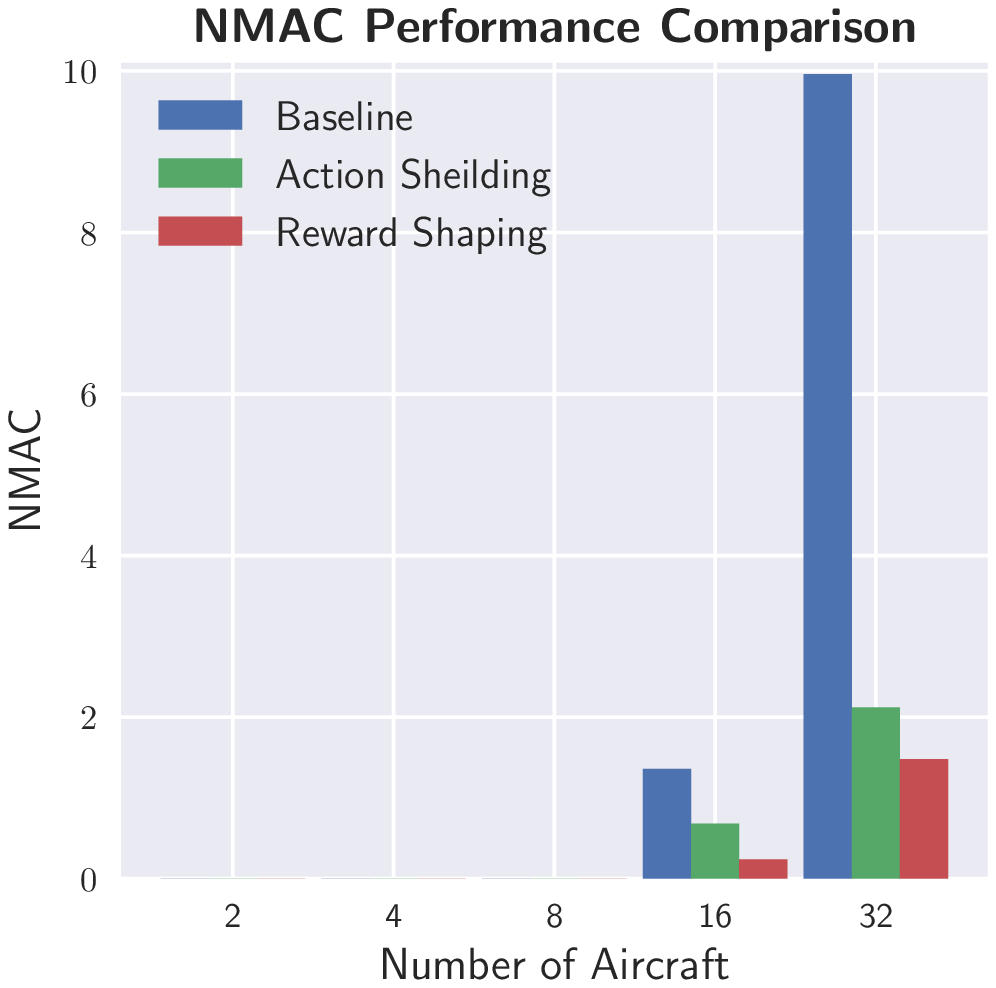}
  \caption{NMAC performance comparison}
  \label{fig:NMAC_bar}
\end{subfigure}
\hspace{2em}
\begin{subfigure}[b]{0.45\textwidth}
  \includegraphics[width=\linewidth]{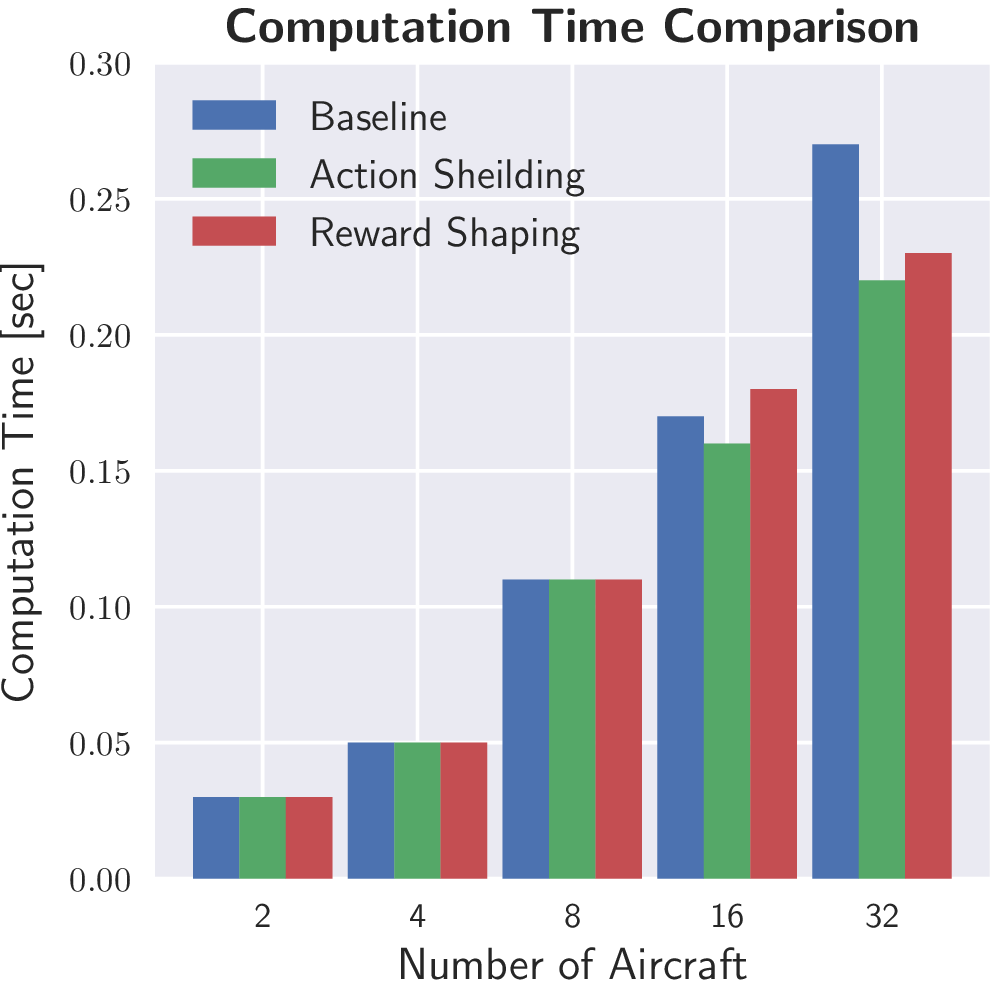}
  \caption{Computation time comparison}
  \label{fig:CompTime_1}
\end{subfigure}
\caption{Performance comparison of the three proposed methods. }
\label{three_comparison}
\end{figure}

A performance comparison of the three proposed methods is shown in Figure \ref{three_comparison}. The results, as depicted in Figure \ref{fig:NMAC_bar}, indicate that the baseline trajectory planner, which does not utilize any reinforcement techniques, exhibits poor safety performance. In contrast, the trajectory planner utilizing reward shaping demonstrates the best performance. While the implementation of action shielding improves the performance of the baseline trajectory planner, it still falls short in comparison to the trajectory planner utilizing reward shaping. Figure \ref{fig:CompTime_1} illustrates the computational time performance comparison of the proposed methods, where it is evident that the differences in performance are minimal.

\subsection{State Constraints}
\begin{figure}[h]
\begin{center}
\includegraphics[width=0.8\textwidth]{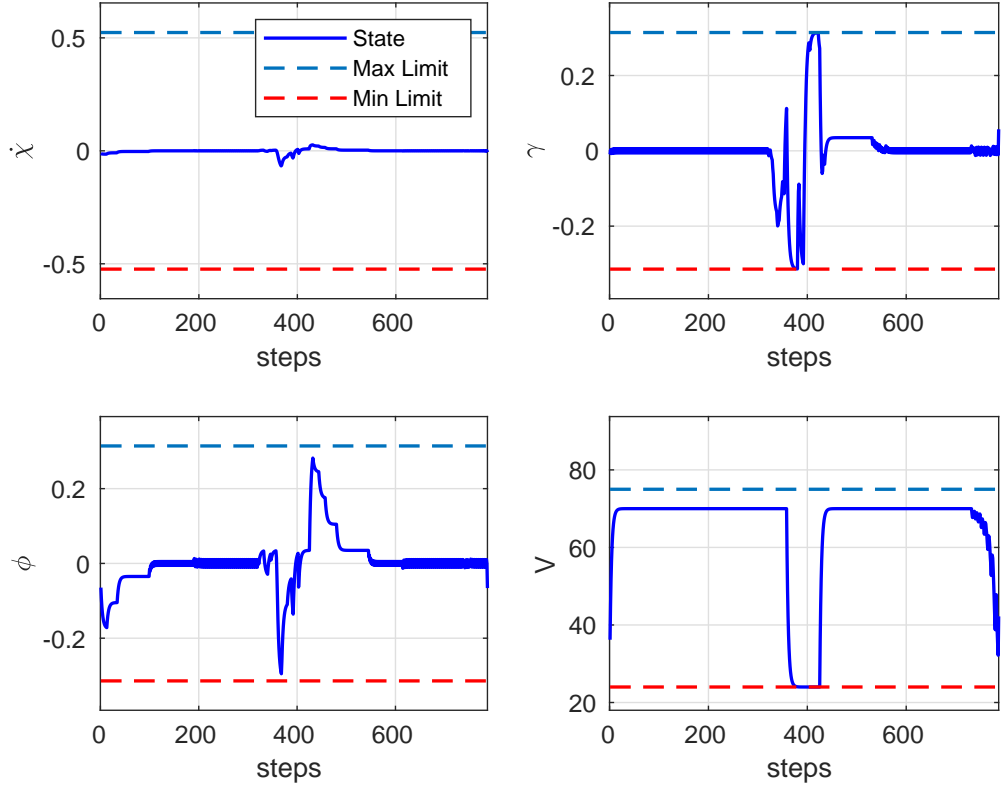}
\caption{\textbf{Constrained state trajectories of an aircraft.}}
\label{fig:state_cons}
\end{center}
\end{figure}

This study also incorporates constraints on certain aircraft states, as presented in Table \ref{state_constraint}, to approximate the operation of air taxis and ensure passenger comfort. These constraints restrict the aircraft from performing maneuvers that may cause discomfort to passengers. Additionally, a constraint on velocity has been imposed to avoid operation below the stall speed of the aircraft. The state trajectories of an aircraft are presented in Fig \ref{fig:state_cons}. It can be observed from the figure that the state trajectories of the aircraft consistently adhered to the imposed constraints throughout the operation of the aircraft.      

\begin{table}[h]
\begin{center}
    \begin{tabular}{ ||c c c c  c c c c|| } 
 \hline 
 \thead{$V_{min}$ \\(Kts)} & \thead{ $V_{max}$ \\(Kts)} & \thead{$\dot{\chi}_{min}$\\(deg/sec) } & \thead{ $\dot{\chi}_{max}$ \\(deg/sec)}& \thead{$\phi_{min}$ \\(deg)}& \thead{ $\phi_{max}$ \\(deg) } & \thead{$\gamma_{min}$ \\(deg)} & \thead{$\gamma_{max}$ \\(deg)} \\  \hline \hline
 47 & 133 & -30 & 30 & -5 & 20 & -20 & 20\\ \hline

\end{tabular}
    \caption{ \textbf{Limits on aircraft performance to approximate an air taxi}}
    \label{state_constraint}
    \end{center}
\end{table}

\section{Conclusion}
\label{sec:conclusion}

This study proposes a safe and scalable trajectory planning framework for urban air mobility (UAM) systems. The proposed framework operates in a decentralized manner, allowing each aircraft to independently plan its trajectory based on information about its surrounding environment. The framework employs a Markov Decision Process (MDP)-based trajectory planner and a data-driven reachability analysis module to synthesize each aircraft's trajectory in real-time. To enhance safety performance, techniques such as reward shaping and action shielding have been explored to be included in the overall framework. The effectiveness of the framework has been evaluated through simulations involving up to 32 aircraft in UAM scenarios, and the results demonstrate the computational efficiency and safe operation of the trajectory planner. Future research will aim to optimize the quality of the generated trajectories, such as reducing flight time and energy consumption.

\section*{Acknowledgments}
This project is partially supported by NASA Grant 80NSSC21M0087 under the NASA System-Wide Safety (SWS) program. 

\bibliography{sample}

\end{document}